\pdfoutput=1

\documentclass[11pt]{article}

\usepackage{ACL2023}

\usepackage{times}
\usepackage{latexsym}
\usepackage{tabularx}
\usepackage{booktabs}
\usepackage{tabularray}
\usepackage{enumitem}
\usepackage{etoolbox}
\UseTblrLibrary{booktabs}
\usepackage[utf8]{inputenc}
\DeclareUnicodeCharacter{0307}{o}

\usepackage[T1]{fontenc}

\usepackage[utf8]{inputenc}

\usepackage{microtype}

\usepackage{inconsolata}
\usepackage{graphicx}

\definecolor{lightergray}{RGB}{230,230,230}
\definecolor{DarkRed}{RGB}{130,25,0}
\definecolor{DarkGreen}{RGB}{30,130,30}
\definecolor{DarkBlue}{RGB}{0,0,250}
\definecolor{purple}{rgb}{0.5,0,1}
\definecolor{dcyan}{rgb}{0.2,0.6,0.5}
\definecolor{darkgreen}{rgb}{0,200,0}
\definecolor{light-gray}{gray}{0.95} 
\definecolor{darkred}{RGB}{200,0,0}
\definecolor{lightgreen}{RGB}{231,255,219}
\definecolor{lightred}{RGB}{252,231,234}
\definecolor{lightyellow}{RGB}{250,253,191}

\definecolor{xa}{RGB}{250,236,236}
\definecolor{xb}{RGB}{233,155,155}
\definecolor{xc}{RGB}{253,248,248}
\definecolor{xd}{RGB}{241,193,193}
\definecolor{xe}{RGB}{233,154,154}
\definecolor{xf}{RGB}{255,255,255}
\definecolor{xg}{RGB}{231,142,142}

\definecolor{aa}{RGB}{250,236,236}
\definecolor{ab}{RGB}{225,183,183}
\definecolor{ac}{RGB}{253,245,245}
\definecolor{ad}{RGB}{242,195,195}
\definecolor{ae}{RGB}{236,165,165}
\definecolor{af}{RGB}{255,255,255}
\definecolor{ag}{RGB}{231,142,142}

\definecolor{ba}{RGB}{249,228,228}
\definecolor{bb}{RGB}{231,143,143}
\definecolor{bc}{RGB}{253,245,245}
\definecolor{bd}{RGB}{236,171,171}
\definecolor{be}{RGB}{226,120,120}
\definecolor{bf}{RGB}{255,255,255}
\definecolor{bg}{RGB}{220,91,91}

\definecolor{ca}{RGB}{174,0,0}
\definecolor{cb}{RGB}{193,88,88}
\definecolor{cc}{RGB}{215,155,155}
\definecolor{cd}{RGB}{226,185,185}
\definecolor{ce}{RGB}{221,169,169}
\definecolor{cf}{RGB}{225,181,181}
\definecolor{cg}{RGB}{228,189,189}

\usepackage{fixltx2e} 

\usepackage{tikz}

\newcommand{\udensdot}[1]{%
    \tikz[baseline=(todotted.base)]{
        \node[inner sep=1pt,outer sep=0pt] (todotted) {#1};
        \draw[densely dotted] (todotted.south west) -- (todotted.south east);
    }%
}%

\newcommand{\udash}[1]{%
    \tikz[baseline=(todotted.base)]{
        \node[inner sep=1pt,outer sep=0pt] (todotted) {#1};
        \draw[dashed] (todotted.south west) -- (todotted.south east);
    }%
}%

\title{Privacy- and Utility-Preserving NLP with Anonymized Data: A case study of Pseudonymization}

\author{Olexandr Yermilov$^1$\thanks{\hspace{5pt}Work done as an intern at Grammarly.}\ , Vipul Raheja$^2$\ , Artem Chernodub$^2$ \\
$^1$Ukrainian Catholic University, Applied Sciences Faculty, \space $^2$Grammarly \\
\texttt{oleksandr.yermilov@ucu.edu.ua,} \\
\texttt{\{vipul.raheja,artem.chernodub\}@grammarly.com}
}

\begin{document}
\maketitle
\begin{abstract}
This work investigates the effectiveness of different pseudonymization techniques, ranging from rule-based substitutions to using pre-trained Large Language Models (LLMs), on a variety of datasets and models used for two widely used NLP tasks: text classification and summarization. Our work provides crucial insights into the gaps between original and anonymized data (focusing on the pseudonymization technique) and model quality and fosters future research into higher-quality anonymization techniques to better balance the trade-offs between data protection and utility preservation. We make our code, pseudonymized datasets, and downstream models publicly available.\footnote{\url{https://github.com/olexandryermilov/privacy-preserving-nlp}} 
\end{abstract}

\section{Introduction}
With the advances in artificial intelligence and data-hungry machine learning systems, privacy and compliant data governance have become increasingly important. Text documents, such as emails, court rulings, customer service chats, interview transcripts, and patient records, frequently contain personally identifiable information (PII), such as mentions of persons, locations, organizations, etc. 
 While the collection and use of text data is necessary for improving products or services, conducting research, or providing personalized recommendations, it has to be done in a safe, responsible and compliant way. 
 
 However, access to text data becomes a challenge where data containing personally identifiable mentions is involved. Although it is a widely accepted notion that no data is truly anonymous and is said to be an unattainable target \cite{rocher2019estimating}, pseudonymization, on the other hand, is recognized by the GDPR as one of the ways (and a requirement) to reduce risks of re-identification of a data subject \cite{EuropeanParliament2016a}. Following \citet{eder-etal-2022-beste}, we define \textit{pseudonymization} as recognizing entities bearing privacy-sensitive information, and their replacement by realistic substitutes.

\begin{table}[t]
\small
\centering
\begin{tabular}{p{0.25\linewidth}|p{0.62\linewidth}}
\toprule
\textbf{Original} & \underline{Sarah} works at \udensdot{The Times} in \udash{London} with \underline{\underline{Rachel}} and \underline{\underline{\underline{David}}}.\\ 
\midrule
\textit{Sanitized} & \underline{\texttt{PERSON\_1}} works at \udensdot{\texttt{ORGANIZATION\_1}} in \udash{\texttt{LOCATION\_1}} with \underline{\underline{PERSON\_2}} and \underline{\underline{\underline{\texttt{PERSON\_3}}}}.\\
\midrule
\textit{Pseudonymized} & \underline{Sophie} works at \udensdot{Manchester Evening} \udensdot{News} in \udash{Manchester} with \underline{\underline{Emma}} and \underline{\underline{\underline{Tom}}}.\\
\bottomrule
\end{tabular}
\caption{While the primary focus of our work is Pseudonymization, we use Sanitization as a baseline for comparison. Different types of underlines correspond to different categories of entities to be pseudonymized.}
\vspace{-0.5cm}
\label{table-deidentification-types}
\end{table}

With the right implementation and safeguards, pseudonymization can be a useful technique for protecting the privacy of individuals while still enabling data-driven technological advances, such as NLP research, enabling researchers to work with sensitive data, while reducing data privacy risks. 
However, there is a risk that quality of texts can often be compromised by techniques such as pseudonymization, which can not only negatively affect downstream NLP tasks and analyses, it can also reduce the utility of anonymized data for other research. 
 It is noteworthy that while privacy and utility-preserving NLP has been a crucial topic in the medical domain, it has been largely overlooked in mainstream NLP research, barring a few recent works (Section \ref{sec:related}). 
 The quality of clinical texts can often be compromised by de-identification. 
Therefore, in this work, we investigate the effectiveness of pseudonymization as a technique for working with NLP models. 
Specifically, we consider three different systems for pseudonymization: 
\begin{enumerate}[noitemsep,topsep=0pt]
    \item \textbf{NER}, which uses named entity recognition (NER) models to detect text spans containing PII, and then uses a knowledge graph to replace the detected spans;
    \item \textbf{Seq2Seq}, which formulates the task of pseudonymization as a sequence-to-sequence (Seq2Seq) transformation, using an encoder-decoder model;
    \item \textbf{LLM}, which leverages the zero-shot and few-shot learning capabilities of large, pre-trained language models (LLMs) for performing the task of pseudonymization.
\end{enumerate}

We then use the aforementioned systems to pseudonymize training datasets for two widely-used NLP tasks: text classification and summarization, and evaluate the performance of models (trained on these pseudonymized datasets) on downstream tasks. Through our analyses, we provide crucial insights into the effectiveness of different pseudonymization techniques for data anonymization, and their effect on downstream NLP tasks, from a privacy and utility perspective. Finally, we make our code, pseudonymized datasets, and downstream models publicly available to foster future research into privacy- and utility-preserving NLP.

\section{Related Work}
\label{sec:related}
Pseudonymization has predominantly been researched in Clinical NLP up until recently, focusing on NLP techniques on how to replace PII such as named entities in medical texts, across different languages. For English medical texts, \citet{sweeney1996replacing} was one of the first pseudonymization systems, followed by numerous works such as \citet{sweeney2005patient, uzuner2007evaluating, neamatullah2008automated, meystre2010automatic, kushida2012strategies, carrell2013hiding, 6410029, Meystre2015, DBLP:journals/jamia/DernoncourtLUS17, 10.1016/j.jbi.2017.05.023, iwendi2020n}. 

The techniques proposed in related works range from simply replacing the detected text spans by a placeholders, pseudonyms or synthetic surrogates using lists, lexical substitution such as synonyms or hypernyms, or knowledge bases \cite{lison-etal-2021-anonymisation, 10.1162/coli_a_00458}. 
Relatedly, techniques such as \textit{C}-sanitize \cite{https://doi.org/10.1002/asi.23363}, \textit{t}-plausibility \cite{10.5555/2423656.2423657}, and more recently, \citet{yue-etal-2021-differential} have proposed frameworks for privacy-aware and -preserving document sanitization and pseudonymization.

While numerous recent works such as the aforementioned ones have investigated the topic of pseudonymization, our work comes closest to \citet{Lampoltshammer2019ImpactOA, Obeid2019ImpactOD, berg-etal-2020-impact, vakili-etal-2022-downstream} and \citet{liu2023deidgpt}, which focus on analyzing different techniques of data anonymization or pseudonymization, and their effect on downstream tasks. However, our work differs from those since they focus on different domains, different tasks, and different techniques.

\begin{table*}
\small
\centering
\begin{tabular}{p{0.14\linewidth}|p{0.25\linewidth}|p{0.08\linewidth}|p{0.07\linewidth} |p{0.07\linewidth}|p{0.07\linewidth}|p{0.12\linewidth}} 
\toprule
\textbf{Task} & \textbf{Dataset name} & \textbf{train size} & \textbf{dev size} & \textbf{test size} & \textbf{domain} & \textbf{metrics}   \\
\midrule
Summarization & CNN/DM \cite{nallapati-etal-2016-abstractive} & 286,817 & 13,368 & 11,487 & news & ROUGE-1/2/L  \\
\midrule
Text classification & IMDB \cite{maas-etal-2011-learning} & 25,000 & N/A & 25,000 & movie reviews & F-score \\
\bottomrule
\end{tabular}
\caption{Details of the evaluated downstream tasks.}
\vspace{-0.2cm}
\label{table-dowstream-tasks}
\end{table*}

\section{Pseudonymization Systems}
The general architecture of a pseudonymization system consists of two steps, where they first recognize entities bearing PII (detection), and the second sub-system their replacement by realistic substitutes (replacement). For this work, we restrict our analysis to three predominant categories of named entities: \texttt{PERSON (PER)}, \texttt{LOCATION (LOC)}, and \texttt{ORGANIZATION (ORG)}. Using this general framework, we describe the three types of systems that are used in our experiments:

\subsection{NER-based Pseudonymization (NER-PS)}\label{sec:ner-pseudo}
The NER-based system uses an off-the-shelf Named Entity Recognition (NER) system to first detect spans of named entities that belong to the aforementioned categories. We use two publicly available NER systems for the first stage: spaCy\footnote{We use spaCy v3.5.1: \url{spacy.io/usage/v3-5}} and FLAIR\footnote{We use FLAIR v0.12.2: \url{github.com/flairNLP/flair}}. The Spacy NER is a fine-tuned RoBERTa model \cite{liu2019roberta}, whereas the FLAIR NER is a LSTM-CRF model based on Flair embeddings \cite{akbik-etal-2018-contextual}. 

The detected named entity spans are then replaced with named entities having similar characteristics, such as gender and language of origin (as described in Wikidata) for \texttt{PER}s, and so on. We first generate a list of replacement candidates, and then randomly sample a single item from this list under some predefined constraints (details in \ref{app:nerps}).

We refer to the two NER-based systems as \textbf{NER-PS\textsubscript{(SPACY)}} and \textbf{NER-PS\textsubscript{(FLAIR)}}.




\subsection{Seq2Seq Pseudonymization (Seq2Seq-PS)}
The Seq2Seq-based system was developed by fine-tuning a BART-\texttt{base} model \cite{lewis-etal-2020-bart} on a parallel corpus of pseudonymized texts (created using the NER-PS system). An important thing to note is that this system does not exactly fit the two-step process outlined above, as it performs the full task in a single-step text-to-text transformation. 

Specifically, we developed two variants of this system using the same NER models as the NER-PS. We refer to the two Seq2Seq-PS variants as 
\textbf{Seq2Seq-PS\textsubscript{(SPACY)}}, \textbf{Seq2Seq-PS\textsubscript{(FLAIR)}}, depending on which NER-PS system was used to create the parallel training data for training the system.

\subsection{LLM Pseudonymization (LLM-PS)}
Following the aforementioned two-step architecture, the LLM-based system is based on a sequential chain of two LLMs: GPT-3\footnote{We use \texttt{text-curie-001} as the GPT-3 model.} \cite{NEURIPS2020_1457c0d6} and ChatGPT\footnote{We use \texttt{gpt-turbo-3.5} as the GPT-3.5 model.}.
For the first step, we extract named entities using GPT-3 with a 1-shot prompt (details in Appendix \ref{app:prompt}), and then perform 1-shot pseudonymization on the extracted named entities using ChatGPT. 
 
We chose GPT-3 to perform the detection step of the architecture due to the fact it works much faster on big paragraphs of text (characterized by both text classification and summarization tasks). Despite being considerably slow, we chose ChatGPT (GPT-3.5) for the replacement step, since the size of the input text is much smaller for the replacement sub-task, and we observed better qualitative performance with this model compared to GPT-3.

\section{Experiments}
In this section, we experimentally evaluate the considered pseudonymization methods. First, we evaluate the negative impact of pseudonymization on the downstream tasks' quality. Next, we compare the privacy preservation quality of different pseudonymization methods. Finally, we evaluate the consistency and privacy-preservation characteristics of pseudonymized texts through a text syntheticity detection experiment.

\subsection{Downstream Tasks Performance}\label{sec:downstream}
Since pseudonymization may introduce additional noise into the processed data, we evaluate the impact of various pseudonymization methods on target dataset quality for the respective downstream tasks. We first pseudonymize the texts for two downstream tasks: Summarization and Text Classification  (Table \ref{table-dowstream-tasks}), using the aforementioned methods, and then train and evaluate the trained models on their respective task-specific metrics.  

For training, we fine-tune the \texttt{bart-base}\footnote{\url{https://huggingface.co/facebook/bart-base}} \cite{lewis-etal-2020-bart} for the Summarization task, and \texttt{bert-base-cased}\footnote{\url{https://huggingface.co/bert-base-cased}} \cite{devlin-etal-2019-bert} for the Text Classification task. In both scenarios, we train the models for three epochs using \textit{AdamW} optimization \cite{loshchilov2017decoupled} with learning rate $\alpha=2*10^{-5}$, and batch size $8$.

For evaluation, as a baseline, we use the quality obtained with the original (non-pseudonymized) texts using the same training process to make sure the difference in metrics is caused only by the difference in the training datasets. Also, as an additional baseline, we compare the results of pseudonymization with two NER-based sanitizations (Table \ref{table-deidentification-types} for reference) denoted by \textbf{{NER-S}\textsubscript{(SPACY)}} and \textbf{{NER-S}\textsubscript{(FLAIR)}}. The sanitization method is the same as NER-PS (Section \ref{sec:ner-pseudo}) except that the detected named entities are replaced with enumerated placeholders, e.g. {\texttt{PERSON\_1}}, {\texttt{LOCATION\_2}}, and {\texttt{ORGANIZATION\_3}}, instead of Wikidata-based named entities.
\begin{table}[t]
\tiny
\begin{booktabs}{
  colspec = {l|ccc|c},
  cell{1}{4} = {c=1}{c}, 
  cell{1}{2} = {c=3}{c} 
}
\toprule
& \textbf{Summarization} & & & \textbf{Classification} \\
\cmidrule[lr]{2-4}\cmidrule[lr]{5-5}
& \textbf{ROUGE-1} & \textbf{ROUGE-2} & \textbf{ROUGE-L} & \textbf{F-score} \\
 \toprule
\textbf{Original text} & \textbf{42.82} &  \textbf{20.13} &  \textbf{36.33} & \textbf{88.42} \\
\midrule
\textbf{{NER-S}\textsubscript{(SPACY)}} &   41.59 &  19.17 & 29.07 & 87.65 \\
\textbf{NER-S\textsubscript{(FLAIR)}} & 39.05 &  17.52 & 27.43 & 87.88 \\
\midrule
\textbf{NER-PS\textsubscript{(SPACY)}} &  \textbf{41.93} &  \textbf{19.38} & \textbf{29.36} & 88.06 \\
\textbf{NER-PS\textsubscript{(FLAIR)}} &  40.25  &  18.04 & 27.97 & 88.14 \\
\midrule
\textbf{S2S-PS\textsubscript{(SPACY)}} & 39.1 & 17.23 & 26.96 & 88.10 \\
\textbf{S2S-PS\textsubscript{(FLAIR)}} & 36.04 & 15.07 &24.73 & 88.13 \\
\midrule
\textbf{LLM-PS} & 38.62 & 16.57 & 26.34 & \textbf{88.15} \\
\bottomrule
\end{booktabs}
\caption{Results of downstream evaluation tasks: summarization (left) and text classification (right). The smaller the gap with the original text, the better the utility is preserved.}
\vspace{-0.5cm}
\label{table-summarization-results}
\end{table}

Evaluation results on both the downstream tasks are presented in Table \ref{table-summarization-results}.
We observe that NER-based pseudonymization achieves the best results for the summarization task, and approaches with spaCy as the underlying NER system show better results compared to FLAIR. These results are related to the fact that FLAIR is a better NER system, which results in making more changes to the original text and introducing more noise into the dataset. This is further compounded with LLM-PS, as it performs a greater amount of edits, thus, forcing the summarization model to learn different patterns than the original dataset, leading to lower ROUGE scores. 

For the classification task, all pseudonymization approaches show similar results, although using FLAIR as the underlying system results in better classification performance compared to spaCy. The difference in task formulations explains this small difference between methods: sentiment classification mostly relies on words with positive/negative sentiment, not on the named entities in the text (although, named entities might associate with positive/negative sentiment more than others \cite{Batra2010EntityBS}, resulting in a correlation between them and sentiment of the text). Hence, pseudonymization might have a very limited effect on the task-specific performance. On the other hand, the summarization task is more sensitive to any errors introduced by the NER/Replacement models, as any false positives or false negatives might lead to inconsistent entity mentions and entity relationships, leading to a corruption in the data, which might be further learned by the summarization model. 

\subsection{Privacy Preservation}
Another risk with pseudonymization is that some named entities will still remain non-anonymized. To estimate these risks of false negatives, we evaluate our methods of pseudonymization on a standard NER benchmark: The English CoNLL-2003 test set \cite{tjong-kim-sang-de-meulder-2003-introduction}. We pseudonymize the dataset, and compare the resulting texts to the originals. We measure the percentage of named entities of each type in the original texts that get leaked into the pseudonymized texts. 

\begin{table}[t]
\centering
\small
\begin{tabular}{l|c|c|c|c}
\toprule
 & \textbf{PER} & \textbf{ORG} & \textbf{LOC} & \textbf{Mean} \\ 
\midrule
\textbf{NER-PS\textsubscript{(SPACY)}} & {23.00} & {37.9} & \textbf{19.48} & {27.23}  \\ 
\textbf{NER-PS\textsubscript{(FLAIR)}} & \textbf{2.48} & \textbf{10.09} & 21.55 & \textbf{10.23}  \\ 
\midrule
\textbf{Seq2Seq-PS\textsubscript{(SPACY)}} & {70.14} & {78.68} & {79.74} & {75.67}  \\ 
\textbf{Seq2Seq-PS\textsubscript{(FLAIR)}} & {14.82} & {36.65} & {65.76} & {36.03}  \\ 
\midrule
\textbf{LLM-PS} & {34.36} & {33.09} & {40.36} & {35.53}  \\ 
\bottomrule
\end{tabular}
\caption{Results of privacy preservation experiment on CoNLL-2003 test set. We report the False Negative Rate for each type of named entity. Lower is better.}
\label{table-privacy-estimation-results}
\end{table}
We observe that NER-based approaches show better results than Seq2Seq approaches, and FLAIR approaches show better results compared to their spaCy equivalents (Table \ref{table-privacy-estimation-results}), which confirms the observations of the previous experiment. Similar to the observations in Section \ref{sec:downstream}, the former observation is related to the fact that the errors present in NER systems are propagated into the Seq2Seq approaches due to the way they were trained. 

\subsection{Text Syntheticity Detection}
As mentioned above, pseudonymization may corrupt relationships and alignment among named entities and other artifacts in the text. 
For example, the United States never had a president named "John Smith." 
Due to such contextual distortions, pseudonymization can negatively affect the quality of processed texts in hard-to-predict ways.

To estimate the degree to which pseudonymized texts are similar to natural ones, we carry out a text syntheticity detection experiment. We combine original and pseudonymized texts from the Summarization task to a single dataset, and train a text  classification model with the goal of detecting pseudonymized texts from their non-pseudonymized counterparts, using the same model and settings as for the Text Classification task (Section \ref{sec:downstream}). The results are presented in Table \ref{table-anonymization-detection-results}. 

LLM-PS shows the best results for this experiment, which are about an order of magnitude better than replacement-based pseudonymization methods. We observe that it is happening because in LLM-rewritten texts, named entities are in better agreement with the context, making it the best-performing system for preserving the syntactic and semantic integrity of the original text.

\begin{table}
\centering
\small
\begin{tabular}{l|c|c|c}
\toprule
 & \textbf{Precision} & \textbf{Recall}&\textbf{F-score} \\ 
\midrule
\textbf{NER-PS\textsubscript{(SPACY)}} & 99.12 & 97.86 & 98.49  \\ 
\textbf{NER-PS\textsubscript{(FLAIR)}} & 98.68 & 95.96& 97.30   \\ 
\midrule
\textbf{Seq2Seq-PS\textsubscript{(SPACY)}}  & 99.94 &  99.76 &  99.85  \\
\textbf{Seq2Seq-PS\textsubscript{(FLAIR)}} & 99.61 & 98.41 & 99.01 \\ 
\midrule
\textbf{LLM-PS} & \textbf{85.61} & \textbf{66.92} & \textbf{75.12} \\ 
\bottomrule
\end{tabular}
\caption{Results of text syntheticity detection experiment. Lower is better.}
\vspace{-0.4cm}
\label{table-anonymization-detection-results}
\end{table}
\section{Conclusions}
We investigate the effectiveness of 
pseudonymization 
for NLP research with privacy-sensitive data. We develop three different approaches for this task, and evaluate them from three aspects: downstream task performance (on two downstream tasks: text summarization and text classification), privacy preservation, and text syntheticity detection. We find that the proposed approaches have pros and cons for pseudonymization, so one must chose what task and objective (privacy vs. utility) is the most important for them. NER-based systems with FLAIR perform the best for privacy preservation and downstream task performance, whereas the LLM-based system shows the best results for preserving the integrity of the text.

\section*{Limitations}
While we endeavor in this work to shed light on the impact of various pseudonymization techniques, we recognize a major limitation of our work -- especially the LLM-based pseudonymization approach.  Using closed-source LLMs may not be an acceptable solution for many settings since it requires sending a (potentially sensitive) text to a third-party API, which, in the absence of appropriate legal 
safeguards and responsible-use agreements, defeats the purpose of privacy preservation. 

There are some more technical limitations of the work, such as the following: 
\begin{itemize}[noitemsep,nolistsep]
    \item While this is a problem that affects sensitive texts in all languages, all the experiments were conducted for data in the English language only. 
    \item LLMs are highly sensitive to prompts, as well as the number and ordering of examples provided for few-shot learning. In this work, we experimented with a limited number of prompts for LLM-PS due to API cost constraints.
    \item For the data privacy detection experiment, the FLAIR NER system was trained using the CoNLL-2003 dataset, which might affect its performance for privacy protection tasks. This may also apply to GPT-3 and ChatGPT models as the authors do not state specifically on which data they were trained.
    \item We considered only a limited part of named entity types, specifically, \texttt{PERSON (PER)}, \texttt{LOCATION (LOC)}, and \texttt{ORGANIZATION (ORG)}, whereas it is well understood that PII encompasses a much broader range of data types (eg. dates, phone numbers, etc.). We also do not consider sentiments associated with named entities used for substitution in the downstream task of text classification. 
\end{itemize}    
We plan to address these in future work.

\section*{Ethics Statement}
User data privacy and data anonymization, are sensitive, and very important matters. Through this work, we try to dive deeper into the challenges and opportunities of using pseudonymization as a technique to strike a suitable tradeoff between privacy- and utility preservation. The goal of this work is to expose the strengths and limitations of different techniques and their implications. The datasets, knowledge bases, and models that we work with have been publicly released for many years. All of these artifacts are considered to be in the public sphere from a privacy perspective. We do not make any recommendations on using these on public or private datasets without proper due diligence for privacy, security, legal, and compliance measures. 

Another risk is that pseudonymization may corrupt the names of people, organizations, and locations and state them in an inappropriate context and therefore produce offensive texts.

\section{Acknowledgements}
We express our gratitude to our colleagues Cortney Napoles and Leonardo Neves for their advice and to our managers Viktor Zamaruiev and Max Gubin for their constant support. To our communities: While we are writing this, our homeland Ukraine continues to resist the unprovoked Russian invasion. We are grateful to everyone who defends Ukraine, declares support for the people of Ukraine, and sends aid. Thank you!

\bibliography{anthology,custom}

\begin{thebibliography}{38}
\expandafter\ifx\csname natexlab\endcsname\relax\def\natexlab#1{#1}\fi

\bibitem[{Akbik et~al.(2018)Akbik, Blythe, and
  Vollgraf}]{akbik-etal-2018-contextual}
Alan Akbik, Duncan Blythe, and Roland Vollgraf. 2018.
\newblock \href {https://aclanthology.org/C18-1139} {Contextual string
  embeddings for sequence labeling}.
\newblock In \emph{Proceedings of the 27th International Conference on
  Computational Linguistics}, pages 1638--1649, Santa Fe, New Mexico, USA.
  Association for Computational Linguistics.

\bibitem[{Anandan et~al.(2012)Anandan, Clifton, Jiang, Murugesan,
  Pastrana-Camacho, and Si}]{10.5555/2423656.2423657}
Balamurugan Anandan, Chris Clifton, Wei Jiang, Mummoorthy Murugesan, Pedro
  Pastrana-Camacho, and Luo Si. 2012.
\newblock T-plausibility: Generalizing words to desensitize text.
\newblock \emph{Trans. Data Privacy}, 5(3):505–534.

\bibitem[{Batra and Rao(2010)}]{Batra2010EntityBS}
Siddharth Batra and D.T.V~Dharmajee Rao. 2010.
\newblock Entity based sentiment analysis on twitter.

\bibitem[{Berg et~al.(2020)Berg, Henriksson, and
  Dalianis}]{berg-etal-2020-impact}
Hanna Berg, Aron Henriksson, and Hercules Dalianis. 2020.
\newblock \href {https://doi.org/10.18653/v1/2020.louhi-1.1} {The impact of
  de-identification on downstream named entity recognition in clinical text}.
\newblock In \emph{Proceedings of the 11th International Workshop on Health
  Text Mining and Information Analysis}, pages 1--11, Online. Association for
  Computational Linguistics.

\bibitem[{Brown et~al.(2020)Brown, Mann, Ryder, Subbiah, Kaplan, Dhariwal,
  Neelakantan, Shyam, Sastry, Askell, Agarwal, Herbert-Voss, Krueger, Henighan,
  Child, Ramesh, Ziegler, Wu, Winter, Hesse, Chen, Sigler, Litwin, Gray, Chess,
  Clark, Berner, McCandlish, Radford, Sutskever, and
  Amodei}]{NEURIPS2020_1457c0d6}
Tom Brown, Benjamin Mann, Nick Ryder, Melanie Subbiah, Jared~D Kaplan, Prafulla
  Dhariwal, Arvind Neelakantan, Pranav Shyam, Girish Sastry, Amanda Askell,
  Sandhini Agarwal, Ariel Herbert-Voss, Gretchen Krueger, Tom Henighan, Rewon
  Child, Aditya Ramesh, Daniel Ziegler, Jeffrey Wu, Clemens Winter, Chris
  Hesse, Mark Chen, Eric Sigler, Mateusz Litwin, Scott Gray, Benjamin Chess,
  Jack Clark, Christopher Berner, Sam McCandlish, Alec Radford, Ilya Sutskever,
  and Dario Amodei. 2020.
\newblock \href
  {https://proceedings.neurips.cc/paper_files/paper/2020/file/1457c0d6bfcb4967418bfb8ac142f64a-Paper.pdf}
  {Language models are few-shot learners}.
\newblock In \emph{Advances in Neural Information Processing Systems},
  volume~33, pages 1877--1901. Curran Associates, Inc.

\bibitem[{Carrell et~al.(2013)Carrell, Malin, Aberdeen, Bayer, Clark, Wellner,
  and Hirschman}]{carrell2013hiding}
David Carrell, Bradley Malin, John Aberdeen, Samuel Bayer, Cheryl Clark, Ben
  Wellner, and Lynette Hirschman. 2013.
\newblock Hiding in plain sight: use of realistic surrogates to reduce exposure
  of protected health information in clinical text.
\newblock \emph{Journal of the American Medical Informatics Association},
  20(2):342--348.

\bibitem[{Dernoncourt et~al.(2017)Dernoncourt, Lee, Uzuner, and
  Szolovits}]{DBLP:journals/jamia/DernoncourtLUS17}
Franck Dernoncourt, Ji~Young Lee, {\"{O}}zlem Uzuner, and Peter Szolovits.
  2017.
\newblock \href {https://doi.org/10.1093/jamia/ocw156} {De-identification of
  patient notes with recurrent neural networks}.
\newblock \emph{J. Am. Medical Informatics Assoc.}, 24(3):596--606.

\bibitem[{Devlin et~al.(2019)Devlin, Chang, Lee, and
  Toutanova}]{devlin-etal-2019-bert}
Jacob Devlin, Ming-Wei Chang, Kenton Lee, and Kristina Toutanova. 2019.
\newblock \href {https://doi.org/10.18653/v1/N19-1423} {{BERT}: Pre-training of
  deep bidirectional transformers for language understanding}.
\newblock In \emph{Proceedings of the 2019 Conference of the North {A}merican
  Chapter of the Association for Computational Linguistics: Human Language
  Technologies, Volume 1 (Long and Short Papers)}, pages 4171--4186,
  Minneapolis, Minnesota. Association for Computational Linguistics.

\bibitem[{Eder et~al.(2022)Eder, Wiegand, Krieg-Holz, and
  Hahn}]{eder-etal-2022-beste}
Elisabeth Eder, Michael Wiegand, Ulrike Krieg-Holz, and Udo Hahn. 2022.
\newblock \href {https://aclanthology.org/2022.lrec-1.79} {{``}beste
  gr{\"u}{\ss}e, maria meyer{''} {---} pseudonymization of privacy-sensitive
  information in emails}.
\newblock In \emph{Proceedings of the Thirteenth Language Resources and
  Evaluation Conference}, pages 741--752, Marseille, France. European Language
  Resources Association.

\bibitem[{{European Commission}(2016)}]{EuropeanParliament2016a}
{European Commission}. 2016.
\newblock \href {https://eur-lex.europa.eu/eli/reg/2016/679/oj} {Regulation
  ({EU}) 2016/679 of the {European} {Parliament} and of the {Council} of 27
  {April} 2016 on the protection of natural persons with regard to the
  processing of personal data and on the free movement of such data, and
  repealing {Directive} 95/46/{EC} ({General} {Data} {Protection} {Regulation})
  ({Text} with {EEA} relevance)}.

\bibitem[{Iwendi et~al.(2020)Iwendi, Moqurrab, Anjum, Khan, Mohan, and
  Srivastava}]{iwendi2020n}
Celestine Iwendi, Syed~Atif Moqurrab, Adeel Anjum, Sangeen Khan, Senthilkumar
  Mohan, and Gautam Srivastava. 2020.
\newblock N-sanitization: A semantic privacy-preserving framework for
  unstructured medical datasets.
\newblock \emph{Computer Communications}, 161:160--171.

\bibitem[{Kushida et~al.(2012)Kushida, Nichols, Jadrnicek, Miller, Walsh, and
  Griffin}]{kushida2012strategies}
Clete~A Kushida, Deborah~A Nichols, Rik Jadrnicek, Ric Miller, James~K Walsh,
  and Kara Griffin. 2012.
\newblock Strategies for de-identification and anonymization of electronic
  health record data for use in multicenter research studies.
\newblock \emph{Medical care}, 50(Suppl):S82.

\bibitem[{Lampoltshammer et~al.(2019)Lampoltshammer, Thurnay, and
  Eibl}]{Lampoltshammer2019ImpactOA}
Thomas~J. Lampoltshammer, L{\"o}rinc Thurnay, and Gregor Eibl. 2019.
\newblock Impact of anonymization on sentiment analysis of twitter postings.
\newblock \emph{Data Science – Analytics and Applications}.

\bibitem[{Lewis et~al.(2020)Lewis, Liu, Goyal, Ghazvininejad, Mohamed, Levy,
  Stoyanov, and Zettlemoyer}]{lewis-etal-2020-bart}
Mike Lewis, Yinhan Liu, Naman Goyal, Marjan Ghazvininejad, Abdelrahman Mohamed,
  Omer Levy, Veselin Stoyanov, and Luke Zettlemoyer. 2020.
\newblock \href {https://doi.org/10.18653/v1/2020.acl-main.703} {{BART}:
  Denoising sequence-to-sequence pre-training for natural language generation,
  translation, and comprehension}.
\newblock In \emph{Proceedings of the 58th Annual Meeting of the Association
  for Computational Linguistics}, pages 7871--7880, Online. Association for
  Computational Linguistics.

\bibitem[{Lison et~al.(2021)Lison, Pil{\'a}n, Sanchez, Batet, and
  {\O}vrelid}]{lison-etal-2021-anonymisation}
Pierre Lison, Ildik{\'o} Pil{\'a}n, David Sanchez, Montserrat Batet, and Lilja
  {\O}vrelid. 2021.
\newblock \href {https://doi.org/10.18653/v1/2021.acl-long.323} {Anonymisation
  models for text data: State of the art, challenges and future directions}.
\newblock In \emph{Proceedings of the 59th Annual Meeting of the Association
  for Computational Linguistics and the 11th International Joint Conference on
  Natural Language Processing (Volume 1: Long Papers)}, pages 4188--4203,
  Online. Association for Computational Linguistics.

\bibitem[{Liu et~al.(2019)Liu, Ott, Goyal, Du, Joshi, Chen, Levy, Lewis,
  Zettlemoyer, and Stoyanov}]{liu2019roberta}
Yinhan Liu, Myle Ott, Naman Goyal, Jingfei Du, Mandar Joshi, Danqi Chen, Omer
  Levy, Mike Lewis, Luke Zettlemoyer, and Veselin Stoyanov. 2019.
\newblock Roberta: A robustly optimized bert pretraining approach.
\newblock \emph{arXiv preprint arXiv:1907.11692}.

\bibitem[{Liu et~al.(2017)Liu, Tang, Wang, and
  Chen}]{10.1016/j.jbi.2017.05.023}
Zengjian Liu, Buzhou Tang, Xiaolong Wang, and Qingcai Chen. 2017.
\newblock \href {https://doi.org/10.1016/j.jbi.2017.05.023} {De-identification
  of clinical notes via recurrent neural network and conditional random field}.
\newblock \emph{J. of Biomedical Informatics}, 75(S):S34–S42.

\bibitem[{Liu et~al.(2023)Liu, Yu, Zhang, Wu, Cao, Dai, Zhao, Liu, Shen, Li,
  Liu, Zhu, and Li}]{liu2023deidgpt}
Zhengliang Liu, Xiaowei Yu, Lu~Zhang, Zihao Wu, Chao Cao, Haixing Dai, Lin
  Zhao, Wei Liu, Dinggang Shen, Quanzheng Li, Tianming Liu, Dajiang Zhu, and
  Xiang Li. 2023.
\newblock \href {http://arxiv.org/abs/2303.11032} {Deid-gpt: Zero-shot medical
  text de-identification by gpt-4}.

\bibitem[{Loshchilov and Hutter(2017)}]{loshchilov2017decoupled}
Ilya Loshchilov and Frank Hutter. 2017.
\newblock Decoupled weight decay regularization.
\newblock \emph{arXiv preprint arXiv:1711.05101}.

\bibitem[{Maas et~al.(2011)Maas, Daly, Pham, Huang, Ng, and
  Potts}]{maas-etal-2011-learning}
Andrew~L. Maas, Raymond~E. Daly, Peter~T. Pham, Dan Huang, Andrew~Y. Ng, and
  Christopher Potts. 2011.
\newblock \href {https://aclanthology.org/P11-1015} {Learning word vectors for
  sentiment analysis}.
\newblock In \emph{Proceedings of the 49th Annual Meeting of the Association
  for Computational Linguistics: Human Language Technologies}, pages 142--150,
  Portland, Oregon, USA. Association for Computational Linguistics.

\bibitem[{Mamede et~al.(2016)Mamede, Baptista, and Dias}]{7743936}
Nuno Mamede, Jorge Baptista, and Francisco Dias. 2016.
\newblock \href {https://doi.org/10.1109/CEC.2016.7743936} {Automated
  anonymization of text documents}.
\newblock In \emph{2016 IEEE Congress on Evolutionary Computation (CEC)}, pages
  1287--1294.

\bibitem[{Meystre(2015)}]{Meystre2015}
Stephane~M. Meystre. 2015.
\newblock \href {https://doi.org/10.1007/978-3-319-23633-9_26}
  {\emph{De-identification of Unstructured Clinical Data for Patient Privacy
  Protection}}, pages 697--716. Springer International Publishing, Cham.

\bibitem[{Meystre et~al.(2010)Meystre, Friedlin, South, Shen, and
  Samore}]{meystre2010automatic}
Stephane~M Meystre, F~Jeffrey Friedlin, Brett~R South, Shuying Shen, and
  Matthew~H Samore. 2010.
\newblock Automatic de-identification of textual documents in the electronic
  health record: a review of recent research.
\newblock \emph{BMC medical research methodology}, 10(1):1--16.

\bibitem[{Nallapati et~al.(2016)Nallapati, Zhou, dos Santos, Gu̇l{\c{c}}ehre,
  and Xiang}]{nallapati-etal-2016-abstractive}
Ramesh Nallapati, Bowen Zhou, Cicero dos Santos, {\c{C}}a{\u{g}}lar
  Gu̇l{\c{c}}ehre, and Bing Xiang. 2016.
\newblock \href {https://doi.org/10.18653/v1/K16-1028} {Abstractive text
  summarization using sequence-to-sequence {RNN}s and beyond}.
\newblock In \emph{Proceedings of the 20th {SIGNLL} Conference on Computational
  Natural Language Learning}, pages 280--290, Berlin, Germany. Association for
  Computational Linguistics.

\bibitem[{Neamatullah et~al.(2008)Neamatullah, Douglass, Lehman, Reisner,
  Villarroel, Long, Szolovits, Moody, Mark, and
  Clifford}]{neamatullah2008automated}
Ishna Neamatullah, Margaret~M Douglass, Li-Wei~H Lehman, Andrew Reisner,
  Mauricio Villarroel, William~J Long, Peter Szolovits, George~B Moody, Roger~G
  Mark, and Gari~D Clifford. 2008.
\newblock Automated de-identification of free-text medical records.
\newblock \emph{BMC medical informatics and decision making}, 8(1):1--17.

\bibitem[{Obeid et~al.(2019)Obeid, Heider, Weeda, Matuskowitz, Carr, Gagnon,
  Crawford, and Meystre}]{Obeid2019ImpactOD}
Jihad~S. Obeid, Paul~M. Heider, Erin~R. Weeda, Andrew~J. Matuskowitz,
  Christine~M. Carr, Kevin Gagnon, Tami~L. Crawford, and St{\'e}phane~M.
  Meystre. 2019.
\newblock Impact of de-identification on clinical text classification using
  traditional and deep learning classifiers.
\newblock \emph{Studies in health technology and informatics}, 264:283 -- 287.

\bibitem[{Papadopoulou et~al.(2022)Papadopoulou, Lison, {\O}vrelid, and
  Pil{\'a}n}]{papadopoulou-etal-2022-bootstrapping}
Anthi Papadopoulou, Pierre Lison, Lilja {\O}vrelid, and Ildik{\'o} Pil{\'a}n.
  2022.
\newblock \href {https://aclanthology.org/2022.lrec-1.476} {Bootstrapping text
  anonymization models with distant supervision}.
\newblock In \emph{Proceedings of the Thirteenth Language Resources and
  Evaluation Conference}, pages 4477--4487, Marseille, France. European
  Language Resources Association.

\bibitem[{Pilán et~al.(2022)Pilán, Lison, Øvrelid, Papadopoulou, Sánchez,
  and Batet}]{10.1162/coli_a_00458}
Ildikó Pilán, Pierre Lison, Lilja Øvrelid, Anthi Papadopoulou, David
  Sánchez, and Montserrat Batet. 2022.
\newblock \href {https://doi.org/10.1162/coli_a_00458} {{The Text Anonymization
  Benchmark (TAB): A Dedicated Corpus and Evaluation Framework for Text
  Anonymization}}.
\newblock \emph{Computational Linguistics}, 48(4):1053--1101.

\bibitem[{Rocher et~al.(2019)Rocher, Hendrickx, and
  De~Montjoye}]{rocher2019estimating}
Luc Rocher, Julien~M Hendrickx, and Yves-Alexandre De~Montjoye. 2019.
\newblock Estimating the success of re-identifications in incomplete datasets
  using generative models.
\newblock \emph{Nature communications}, 10(1):1--9.

\bibitem[{Sweeney et~al.(2005)Sweeney, Portell, Houck, Smith, and
  Mentel}]{sweeney2005patient}
James~P Sweeney, Keith~S Portell, James~A Houck, Reginald~D Smith, and John~J
  Mentel. 2005.
\newblock Patient note deidentification using a find-and-replace iterative
  process.
\newblock \emph{Journal of Healthcare Information Management: JHIM},
  19(3):65--70.

\bibitem[{Sweeney(1996)}]{sweeney1996replacing}
Latanya Sweeney. 1996.
\newblock Replacing personally-identifying information in medical records, the
  scrub system.
\newblock In \emph{Proceedings of the AMIA annual fall symposium}, page 333.
  American Medical Informatics Association.

\bibitem[{Sánchez and Batet(2016)}]{https://doi.org/10.1002/asi.23363}
David Sánchez and Montserrat Batet. 2016.
\newblock \href {https://doi.org/https://doi.org/10.1002/asi.23363}
  {C-sanitized: A privacy model for document redaction and sanitization}.
\newblock \emph{Journal of the Association for Information Science and
  Technology}, 67(1):148--163.

\bibitem[{Sánchez et~al.(2013)Sánchez, Batet, and Viejo}]{6410029}
David Sánchez, Montserrat Batet, and Alexandre Viejo. 2013.
\newblock \href {https://doi.org/10.1109/TIFS.2013.2239641} {Automatic
  general-purpose sanitization of textual documents}.
\newblock \emph{IEEE Transactions on Information Forensics and Security},
  8(6):853--862.

\bibitem[{Tjong Kim~Sang and
  De~Meulder(2003)}]{tjong-kim-sang-de-meulder-2003-introduction}
Erik~F. Tjong Kim~Sang and Fien De~Meulder. 2003.
\newblock \href {https://aclanthology.org/W03-0419} {Introduction to the
  {C}o{NLL}-2003 shared task: Language-independent named entity recognition}.
\newblock In \emph{Proceedings of the Seventh Conference on Natural Language
  Learning at {HLT}-{NAACL} 2003}, pages 142--147.

\bibitem[{Uzuner et~al.(2007)Uzuner, Luo, and Szolovits}]{uzuner2007evaluating}
{\"O}zlem Uzuner, Yuan Luo, and Peter Szolovits. 2007.
\newblock Evaluating the state-of-the-art in automatic de-identification.
\newblock \emph{Journal of the American Medical Informatics Association},
  14(5):550--563.

\bibitem[{Vakili et~al.(2022)Vakili, Lamproudis, Henriksson, and
  Dalianis}]{vakili-etal-2022-downstream}
Thomas Vakili, Anastasios Lamproudis, Aron Henriksson, and Hercules Dalianis.
  2022.
\newblock \href {https://aclanthology.org/2022.lrec-1.451} {Downstream task
  performance of {BERT} models pre-trained using automatically de-identified
  clinical data}.
\newblock In \emph{Proceedings of the Thirteenth Language Resources and
  Evaluation Conference}, pages 4245--4252, Marseille, France. European
  Language Resources Association.

\bibitem[{Vrande\v{c}i\'{c}(2012)}]{10.1145/2187980.2188242}
Denny Vrande\v{c}i\'{c}. 2012.
\newblock \href {https://doi.org/10.1145/2187980.2188242} {Wikidata: A new
  platform for collaborative data collection}.
\newblock In \emph{Proceedings of the 21st International Conference on World
  Wide Web}, WWW '12 Companion, page 1063–1064, New York, NY, USA.
  Association for Computing Machinery.

\bibitem[{Yue et~al.(2021)Yue, Du, Wang, Li, Sun, and
  Chow}]{yue-etal-2021-differential}
Xiang Yue, Minxin Du, Tianhao Wang, Yaliang Li, Huan Sun, and Sherman S.~M.
  Chow. 2021.
\newblock \href {https://doi.org/10.18653/v1/2021.findings-acl.337}
  {Differential privacy for text analytics via natural text sanitization}.
\newblock In \emph{Findings of the Association for Computational Linguistics:
  ACL-IJCNLP 2021}, pages 3853--3866, Online. Association for Computational
  Linguistics.

\end{thebibliography}
\bibliographystyle{acl_natbib}

\appendix
\section{Training Details}

\subsection{NER Pseudonymization (NER-PS)}
\label{app:nerps}
As part of the two-step pseudonymization pipeline, for both \textbf{NER-PS\textsubscript{(SPACY)}} and \textbf{NER-PS\textsubscript{(FLAIR)}} systems, we leverage Wikidata for the second step -- generation of candidates for replacement. 

 Following some prior works \cite{7743936, papadopoulou-etal-2022-bootstrapping}, we sample the replacements candidates from Wikidata\footnote{\url{https://www.wikidata.org/wiki/Wikidata:Main\_Page}} \cite{10.1145/2187980.2188242}, a knowledge graph where \textit{objects} (entities) are linked together by \textit{properties}. 
 We consider specific membership properties, namely \texttt{instance\_of} (P31), \texttt{subclass\_of} (P279), and \texttt{part\_of} (P361), indicating a hierarchical association from specific to general. 
 
 Given an entity mention that needs to be replaced, we first find a leaf node in the graph that matches the given entity mention. Then, we traverse the graph to extract sibling nodes via the hierarchical associations, and generate replacement candidates based on additional filters. For instance, we filter \texttt{PERSON} entity candidates with ones that have the same gender and language of origin. For \texttt{ORGANIZATION} entities, similar industry and country; and for \texttt{LOCATION} entities, similar location type and country. We then random sample a single item from this list of filtered candidates under the aforementioned constraints.

\subsection{Seq2Seq Pseudonymization (Seq2Seq-PS)}

We fine-tune \texttt{bart-base}\footnote{\url{https://huggingface.co/facebook/bart-base}} \cite{lewis-etal-2020-bart} for Seq2Seq models. We train the models for three epochs using \textit{AdamW} optimization \cite{loshchilov2017decoupled} with the learning rate $\alpha=2*10^{-5}$, the batch size is $8$.
Training corpus was sampled from the Wikipedia articles and has size of 19M samples.

\subsection{LLM Pseudonymization (LLM-PS)}
\label{app:prompt}
Table \ref{tab:prompts} shows the prompts we have used for calls to GPT-3 and ChatGPT models. In the first prompt, we are giving the example of extracting named entities (specifically, persons, organizations, and locations) from a small paragraph of text. In the second prompt, we are giving the task as a system message and give examples of changing named entities (again, persons, organizations and location) to named entities of the same type. These prompts can be extended to include named entities of other types.
\begin{table*}[ht]
\small
\centering
\begin{tabular}{p{0.1\linewidth}|p{0.1\linewidth}|p{0.6\linewidth}}
\toprule
\textbf{Stage} & \textbf{Model} & \textbf{Illustrative Prompt(s) / API calls} \\
\midrule
NER  & GPT-3 & \texttt{Find all the locations, names and organizations in the following text. Write them separated by commas:}\\
   & & \texttt{\textbf{Text}: Daniel worked in Google for five years before moving from America to France. Daniel is now working with Emma in Danone and living in Paris.}\\
   & & \texttt{\textbf{Answer}: Daniel, Google, America, France, Emma, Danone, Paris.}\\
   & & \texttt{\textbf{Text}: <text-to-anonymize>}\\
   & & \texttt{\textbf{Answer}: <response-from-API>}\\
\midrule
Replacement & ChatGPT & \texttt{\{} \\
    & & \hspace{5pt} \texttt{"role": "system", }\\
    & & \hspace{5pt} \texttt{"content": "Change following named entities using different} \\
    & & \hspace{5pt} \texttt{named entities of the same type."} \\
    & & \texttt{\}, }\\
    & & \texttt{\{} \\
    & & \hspace{5pt} \texttt{"role": "user", }\\ 
    & & \hspace{5pt} \texttt{"content": "Africa, James Potter, Google, Poland, Lily Jameson, Danone"}\\
    & & \texttt{\}, }\\
    & & \texttt{\{} \\
    & & \hspace{5pt} \texttt{"role": "assistant",}\\ 
    & & \hspace{5pt} \texttt{"content": "Asia, John Lennon, Microsoft, Germany, Anna Smith, Starbucks"},\\
    & & \texttt{\}, }\\
    & & \texttt{\{} \\
    & & \hspace{5pt} \texttt{"role": "user", }\\ 
    & & \hspace{5pt} \texttt{"content": <entities-to-pseudonymize>}\\
    & & \texttt{\}, }\\
    & & \texttt{\{} \\
    & & \hspace{5pt} \texttt{"role": "assistant",}\\ 
    & & \hspace{5pt} \texttt{"content": <response-from-API>},\\
    & & \texttt{\}, }\\
\bottomrule
\end{tabular}
\caption{Illustrative prompts for single-shot named entity recognition and replacement tasks for the LLM-PS System.}
\label{tab:prompts}
\end{table*}
However, this approach should be taken with appropriate caution, as it can also change other parts of the text since single-shot GPT-3 might treat other words in the text as named entities. For example, in sample 11165 from IMDB train set, this is the named entities GPT-3 parse from the text: \texttt{Friday the 13th, Bernie, old man, family, Slashers} and here is the pseudonymized response from ChatGPT: \texttt{Halloween, Nancy, young woman, relatives, Killers}. As we can observe, parts of the request which are not named entities changed in a completely different way: \texttt{"family"} was changed to a synonym word \texttt{"relatives"}, while \texttt{"old man"} was changed to an antonym \texttt{"young woman"}.

\section{Data Examples}
\begin{table*}[t]
\tiny
\centering
\begin{tabular}{p{0.1\linewidth}|p{0.3\linewidth}||p{0.5\linewidth} }
\toprule
& \textbf{Text Classification} & \textbf{Text Summarization} \\
\midrule
\textbf{Original} & Does it get any uglier than this? The only good thing in this movie was \textcolor{red}{Natassia Malthe}, with her stunning Norwegian beauty. God, I wish \textcolor{teal}{Michael Ironside} and the \textcolor{purple}{DeLuise} brothers would stop accepting dumb roles in dumb movies! I mean, at least \textcolor{olive}{SeaQuest} was nice! I know Mr. \textcolor{orange}{Ironside} from a lot of movies, he has acted in 164 movies at this date!! It's true that he was rarely in a major role, but still! &  By . \textcolor{red}{Chris Waugh} . \textcolor{teal}{Pep Guardiola} will never be sacked as \textcolor{purple}{Bayern Munich} head coach according to the Bundesliga champions' chairman. \textcolor{olive}{Karl-Heinz Rummenigge} was questioned about whether or not he was worried that many of \textcolor{purple}{Bayern}'s German World Cup-winning stars had yet to return to pre-season training when he made the claim. German newspaper \textcolor{orange}{Welt am Sonntag} carried an article on Sunday claiming \textcolor{teal}{Guardiola}'s side could struggle this season due to the tiring World Cup campaign. VIDEO Scroll down to watch \textcolor{teal}{Pep Guardiola} lose it with a journalist and get soaked in beer.\\
& & \textcolor{purple}{Bayern Munich} chairman \textcolor{olive}{Karl-Heinz Rummenigge} says the club will 'never' sack boss \textcolor{teal}{Pep Guardiola}.
\\ 
\midrule
\textbf{{NER-PS}\textsubscript{(SPACY)}} &
  Does it get any uglier than this? The only good thing in this movie was \textcolor{red}{Boeing Gap}, with her stunning Norwegian beauty. God, I wish \textcolor{teal}{Lakshmi Kevin} and the \textcolor{purple}{Hector} brothers would stop accepting dumb roles in dumb movies! I mean, at least \textcolor{olive}{EGL} was nice! I know Mr. \textcolor{orange}{Dani} from a lot of movies, he has acted in 164 movies at this date!! It's true that he was rarely in a major role, but still! & By . \textcolor{red}{Nikki Scott} . \textit{Pep Guardiola} will never be sacked as \textcolor{purple}{ASV Cham Engelbert Strauss} head coach according to the Bundesliga champions' chairman. \textcolor{olive}{KunzKuppuswamyMarkus Kul} was questioned about whether or not he was worried that many of \textcolor{purple}{ASV Cham Engelbert Strauss}'s German World Cup-winning stars had yet to return to pre-season training when he made the claim. German newspaper \textcolor{orange}{Modernine TV Hub Omnicare} carried an article on Sunday claiming \textcolor{teal}{Gentek}'s side could struggle this season due to the tiring World Cup campaign. VIDEO Scroll down to watch \textit{Pep} \textcolor{teal}{Gentek} lose it with a journalist and get soaked in beer. \\
& &\textcolor{purple}{ASV Cham Engelbert Strauss} chairman \textcolor{olive}{KunzKuppuswamyMarkus Kul} says the club will 'never' sack boss \textcolor{teal}{Xavier Gentek}. \\
\midrule
\textbf{NER-PS\textsubscript{(FLAIR)}} & Does it get any uglier than this? The only good thing in this movie was \textcolor{red}{Delcine Fleak}, with her stunning Norwegian beauty. \textit{Elmore}, I wish \textcolor{teal}{Nicolas Loveridge} and the \textcolor{purple}{Perreira} brothers would stop accepting dumb roles in dumb movies! I mean, at least \textit{SeaQuest} was nice! I know Mr. \textcolor{orange}{Catala} from a lot of movies, he has acted in 164 movies at this date!! It's true that he was rarely in a major role, but still! & By . \textcolor{red}{Robin Kloss} . \textcolor{teal}{Jesús Lascurain} will never be sacked as \textcolor{purple}{BSV Kickers Emden} head coach according to the Bundesliga champions' chairman. \textcolor{olive}{Peyush Herwig} was questioned about whether or not he was worried that many of \textit{\textcolor{purple}{Duchy of Saxe-Weimar-Eisenach}'s} German World Cup-winning stars had yet to return to pre-season training when he made the claim. German newspaper \textcolor{orange}{Der Angriff} carried an article on Sunday claiming \textit{\textcolor{teal}{Lascurain}'s} side could struggle this season due to the tiring World Cup campaign. VIDEO Scroll down to watch \textcolor{teal}{Jesús Lascurain} lose it with a journalist and get soaked in beer.\\& & \textcolor{purple}{BSV Kickers Emden} chairman \textcolor{olive}{Peyush Herwig} says the club will 'never' sack boss \textcolor{teal}{Jesús Lascurain}\\
\midrule
\textbf{S2S-PS\textsubscript{(SPACY)}} &   Does it get any uglier than this? The only good thing in this movie was \textit{Natassia Malthe}, with her stunning Norwegian beauty. God, I wish \textcolor{teal}{Alistair D’Alessandro} and the \textit{DeLuise} brothers would stop accepting dumb roles in dumb movies! I mean, at least \textit{SeaQuest} was nice! I know Mr. \textcolor{orange}{Suryanarayan} from a lot of movies, he has acted in 164 movies at this date!! It's true that he was rarely in a major role, but still. & By . \textcolor{red}{Floor Blythe} . \textit{Pep Guardiola} will never be sacked as \textit{Bayern Munich} head coach according to the Bundesliga champions' chairman. \textit{Karl-Heinz Rummenigge} was questioned about whether or not he was worried that many of \textit{Bayern}'s German World Cup-winning stars had yet to return to pre-season training when he made the claim. German newspaper \textit{Welt am Sonntag} carried an article on Sunday claiming \textit{Guardiola}'s side could struggle this season due to the tiring World Cup campaign. VIDEO Scroll down to watch \textit{Pep Guardiola} lose it with a journalist and get soaked in beer. \\ 
& & \textit{Bayern Munich} chairman \textcolor{olive}{Jörn}\textit{-Heinz Rummenigge} says the club will 'never' sack boss \textit{Pep Guardiola}.\\
\midrule
\textbf{S2S-PS\textsubscript{(FLAIR)}} &   Does it get any uglier than this? The only good thing in this movie was \textcolor{red}{Jyotirmoye Dhanraj}, with her stunning Norwegian beauty. God, I wish \textcolor{teal}{Alvan Kostas} and the \textcolor{purple}{Sivaramakrishna} brothers would stop accepting dumb roles in dumb movies! I mean, at least \textit{SeaQuest} was nice! I know Mr. \textcolor{orange}{Sankar} from a lot of movies, he has acted in 164 movies at this date!! It's true that he was rarely in a major role, but still! & By . \textcolor{red}{Helge Kowalczyk} . \textcolor{teal}{Raghuvinder Cárdenas} will never be sacked as \textcolor{purple}{TSV Heiligheim} head coach according to the Bundesliga champions' chairman. \textcolor{olive}{Gertrudin Günther} was questioned about whether or not he was worried that many of \textit{\textcolor{purple}{SV Altenburg}}'s German World Cup-winning stars had yet to return to pre-season training when he made the claim.
German newspaper \textit{Welt am Sonntag} carried an article on Sunday claiming \textcolor{teal}{Cárdenas}'s side could struggle this season due to the tiring World Cup campaign.
 VIDEO Scroll down to watch \textcolor{teal}{\textit{Gijsbertus} Cárdenas} lose it with a journalist and get soaked in beer . \\ 
& & \textcolor{purple}{TSV  \textit{Heiligenburg}} chairman \textcolor{olive}{Gertrudin \textit{Schleicher}} says the club will 'never' sack boss \textcolor{teal}{\textit{Gijsbertus} Cárdenas}.\\
\midrule
\textbf{LLM-PS} &  Does it get any uglier than this? The only good thing in this movie was \textcolor{red}{Maria Olsen}, with her stunning Norwegian beauty. God, I wish \textcolor{teal}{Tricia Helfer} and the \textcolor{purple}{Hemsworth} brothers would stop accepting dumb roles in dumb movies! I mean, at least \textcolor{olive}{Battlestar Galactica} was nice! I know Mr. \textit{Ironside} from a lot of movies, he has acted in 164 movies at this date !! It's true that he was rarely in a major role, but still! &  By. \textcolor{red}{Amanda Wilson}. \textcolor{teal}{Jürgen Klopp} will never be sacked as \textcolor{purple}{Borussia Dortmund} head coach according to the Bundesliga champions' chairman. \textcolor{olive}{Franz Beckenbauer} was questioned about whether or not he was worried that many of \textit{Bayern}'s UEFA Champions League-winning stars had yet to return to off-season preparations when he made the claim . German newspaper \textit{Welt am Sonntag} carried an article on Sunday claiming \textcolor{teal}{Klopp}'s side could struggle this season due to the tiring World Cup campaign. VIDEO Scroll down to watch \textcolor{teal}{Jürgen Klopp} lose it with a journalist and get soaked in beer. \\ 
& & \textcolor{purple}{Borussia Dortmund} chairman \textcolor{olive}{Franz Beckenbauer} says the club will 'never' sack boss \textcolor{teal}{Jürgen Klopp}. 
\\
\bottomrule
\end{tabular}
\caption{Examples of Pseudonymization}
\label{appendix_examples_imdb}
\end{table*}

\label{app:examples}
Table \ref{appendix_examples_imdb} shows examples of pseudonymization parts of different samples. We can notice the poor performance of S2S-PS\textsubscript{(SPACY)} and preservation of context in LLM-PS generated text.

\end{document}